\RequirePackage{fix-cm}
\documentclass[smallextended]{svjour3}       
\smartqed  
\usepackage{graphicx}
\usepackage[font=small,labelfont=bf]{caption}
\usepackage[font=footnotesize]{subfig}
\graphicspath{{figures/}}
\usepackage{lscape} 
\usepackage{amsmath}
\DeclareMathOperator*{\argminA}{arg\,min} 
\usepackage{multirow}
\usepackage{chngcntr}
\usepackage[sort&compress,numbers]{natbib}

\begin{document}

\title{Generating Unrestricted Adversarial Examples via Three Parameteres
}
\subtitle{Do you have a subtitle?\\ If so, write it here}


\author {Hanieh Naderi         \and
         Leili Goli            \and
         Shohreh Kasaei 
}


\institute{Department of Computer Science and Engineering, Sharif University of Technology \at
              Tehran, Iran \\
              Tel.: +989126846377\\
              \email{hnaderi@ce.sharif.edu}           
           \and
           Department of Computer Science and Engineering, Sharif University of Technology \at
              Tehran, Iran
}

\date{Received: date / Accepted: date}

\maketitle

\setlength\parindent{0pt}. 

\begin{abstract}
Deep neural networks have been shown to be vulnerable to adversarial examples deliberately constructed to misclassify victim models. As most adversarial examples have restricted their perturbations to $L_{p}$-norm, existing defense methods have focused on these types of perturbations and less attention has been paid to unrestricted adversarial examples; which can create more realistic attacks, able to deceive models without affecting human predictions. To address this problem, the proposed adversarial attack generates an unrestricted adversarial example with a limited number of parameters. The attack selects three points on the input image and based on their locations transforms the image into an adversarial example. By limiting the range of movement and location of these three points and using a discriminatory network, the proposed unrestricted adversarial example preserves the image appearance.
Experimental results show that the proposed adversarial examples obtain an average success rate of 93.5\% in terms of human evaluation on the MNIST and SVHN datasets. 
It also reduces the model accuracy by an average of 73\% on six datasets MNIST, FMNIST, SVHN, CIFAR10, CIFAR100, and ImageNet. It should be noted that, in the case of attacks, lower accuracy in the victim model denotes a more successful attack. The adversarial train of the attack also improves model robustness against a randomly transformed image.

\keywords{Unrestricted adversarial examples \and transformation \and attack \and  Adversarial training}
\end{abstract}


\section{Introduction}
\label{intro}
Recently, Deep Neural Networks (DNNs) are migrating from the research world to the real world applications. Despite having high accuracy in different applications, DNNs are vulnerable to Adversarial Examples (AEs). For the first time in 2014, Szegedy et al. published a paper entitled "intriguing  properties  of  neural  network"\cite{szegedy2013intriguing}. It was demonstrated in this paper that by adding small perturbations to the image pixels, the model can be deceived into seeing another image. By adding these perturbations to the image, AEs are generated that look like the input image from the human point of view, but the model misclassifies them with high confidence.
The presence of AEs can cause serious problems. For instance, misclassification can cause catastrophic accidents in self-driving cars, failure in detecting cancer in medical images or affect security systems. 
Most existing methods generate an AE by changing the intensity of image pixels with $L_{p}$-bounded criteria, called norm-constrained AEs \cite{moosavi2016deepfool,szegedy2013intriguing,goodfellow2014explaining,madry2017towards,tramer2017ensemble,kurakin2016adversarial_ML_at_scale, madry2017towards, dong2018boosting,moosavi2016deepfool, carlini2017towards}. Unrestricted Adversarial Examples (UAEs) are another form of adversarial attacks that were first intorduced in \cite{brown2018unrestricted}, which is not limited to the $L_{p}$-norm. Compared to norm-constrained AEs, research on UAEs is comparatively narrow. These kinds of adversarial attacks are more likely to occur in the real world and can be more destructive by directly challenging many practical applications such as face recognition, autonomous driving, and so forth. Therefore, recently the interest in generating UAEs has increased. There are methods that apply spatial deformation through small rotation, translation, shearing, and scaling to convert input images to UAEs \cite{moosavi2018geometric,engstrom2017rotation,2015manitest}. Using natural transformation (such as changing viewpoint, lighting, coloring, etc) \cite{alcorn2019strike} and adding a small patch to original images \cite{eykholt2018robust} are other methods that have been proposed to construct UAEs.
 Adversarial training, known as the best defense method so far \cite{athalye2018obfuscated,tramer2020adaptive}, is a commonly-used defense that directly exploits a given attack in the process of making a model robust to AEs. Therefore, knowing and understanding different attack types is essential for creating effective defense methods. State-of-the-art defense techniques are often created using norm-constrained AEs \cite{tramer2020adaptive}. But, these defenses easily fail against UAEs. This is due to the fact that norms become  in measuring the model robustness against UAEs. Ineffectiveness of many defense techniques against UAEs has made them major obstacles in robustness of deep learning models and has led to increasing interest in constructing different UAEs.
 Adversarial training, known as the best defense method so far \cite{athalye2018obfuscated,tramer2020adaptive}, is a commonly-used defense that directly exploits a given attack in the process of making a model robust to AEs. Therefore, knowing and understanding different attack types is essential for creating effective defense methods. State-of-the-art defense techniques are often created using norm-constrained AEs \cite{tramer2020adaptive}. But, these defenses easily fail against UAEs. This is due to the fact that $L_{p}$-norms become ineffective in measuring the model robustness against UAEs. Ineffectiveness of many defense techniques against UAEs has made them major obstacles in robustness of deep learning models and has led to increasing interest in constructing different UAEs.
In an independent line of research, authors had focused on improving invariant models to transformations by adding modules or modifying model layers \cite{cohen2016group, naderi2020scale, marcos2017rotation, jaderberg2015SpatialTransformerNetworks, laptev2016ti, dai2017deformable,shen2016transform}. The goal in this line of research is to modify the model in a way that applying random transformations to model's inputs would not affect the predictions made by the model. In this paper, the focus is on improving adversarial robustness to images deformed by smallest possible transformation. 

Attacks can be studied through three different aspects. 
\begin{itemize}
\item Poisoning vs evasion attacks, when the former occurs during the training phase and the latter occurs during the testing phase. 
\item Targeted (the model assigns the image class to a target class ) vs untargeted attack (the model assigns the image class to any class other than the correct class). 
\item White-box (the adversary has access to all model information; such as architecture, parameters, and hyperparameters) vs black-box attacks (the adversary only has access to the model output 
\end{itemize}

\begin{figure*}
    \centering
\captionsetup[subfigure]{labelformat=empty}
\captionsetup[subfloat]{position=bottom}
\captionsetup[subfigure]{font=scriptsize,labelfont=small}

 \caption*{\ \ \ \ \  Original \ \ \ \ \ \      Adversary  \ \ \ \ \ \ \ \ \ \    Original  \ \ \ \ \ \     Adversary  \ \ \ \ \ \ \ \ \ \     Original    \ \ \ \ \ \     Adversary  \ \ \  \ }

    \subfloat[8]{{\includegraphics[width=1.6cm]{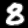} }}
    \subfloat[2]{{\includegraphics[width=1.6cm]{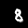} }}
     \qquad
    \subfloat[Chihuahua]{{\includegraphics[width=1.6cm]{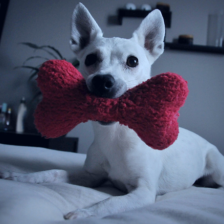} }}
    \subfloat[Seatelt]{{\includegraphics[width=1.6cm]{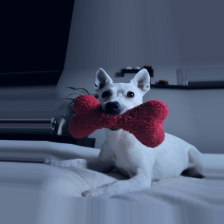} }}
     \qquad
     \subfloat[Pullover]{{\includegraphics[width=1.6cm]{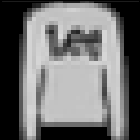} }}
     \subfloat[Bag]{{\includegraphics[width=1.6cm]{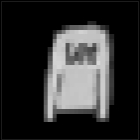} }}
     \qquad

 \caption{Examples of successful urestricted adversarial examples created with proposed attack for MNIST, ImageNet, and FMNIST datasets on LeNet, VGG-16, and ResNet-18 models, respectively. [Odd columns: Original images with labels. Even columns: Adversarial examples with estimated labels.]}
  \label{fig:1}       
\end{figure*}

In this paper, the evasion, untargeted, and black-box attacks are exclusively concerned, although the attack can be easily adapted to targeted settings. A new method of constructing UAEs is presented that can create a combination of natural deformations that can occur in the real world, such as rotation, shift, shear, scale or changing the angle of view in the camera.
The proposed attack can estimate all such deformations when optimized with only three parameters (such examples can be found in Figure \ref{fig:1}). The attack selects three points on the original image and based on their location transforms the image into a UAE. By limiting the amount of movement and location of these three points and using a discriminatory network, the proposed UAE preserves the image appearance.

The rest of the paper is organized as follows. Section 2, summarizes the related work. The proposed UAEs is introduced in Section 3. Experimental results are reported in Section 4.  In Section 5, concluding remarks are discussed.

Contributions of this work are as follows:
\begin{itemize}
\item The UAE is introduced to model natural events that occur in the real-world. This can consistently cause misclassification in a DNN-based classifier under a range of dynamic real conditions, including different viewpoint angles and distances.
\item To have a more accurate model of real-world image distortions where the distortion in scale of the image is not necessarily consistent over the whole image, the proposed method divides the image into smaller parts and distorts the image differently in each part. The special case of having three divisions is thoroughly discussed.
\item Generating UAEs based on spatial deformations by using only three trainable parameters is proposed. The attack method is parameterized with three adjustable parameters and then optimized for these parameters.
\item Robustness to natural perturbations plays a more important role in crafting the models used in real-world scenarios. The proposed attack demonstrates the lack of robustness to such perturbations in currently available state-of-the-art models.
\item Performing adversarial training using the proposed attack can improve robustness against a randomly transformed image.
\item The transferablity of proposed attack is tested on different models over two datasets of CIFAR10 and ImageNet. This analysis shows that our attack can be successfully transferred to different model architectures. 
\end{itemize}

\section{Related Work}
\label{sec:Related Work}

Related work can be divided into two categories. The first category includes methods that construct norm-constrained AEs and the second category includes methods that ignore norm-constrained conditions. 

\subsection{Norm-Constrained Adversarial Examples }
\label{sec:3.1}

Most existing methods try to construct AEs by changing the pixel intensity. Suppose $x$ is the original image and the perturbation $\eta$ is added to it (usually restricted by $L_{p}-norm$ to be small and imperceptible to human). These perturbations start changing image pixel values to the point where they cause a model to make a wrong prediction to craft an AE $x_{adv}=x+||\eta||_{p}$. There are various algorithms for constructing norm-constrained AEs in different settings \cite{goodfellow2014explaining,moosavi2016deepfool,szegedy2013intriguing, madry2017towards,tramer2017ensemble,kurakin2016adversarial_ML_at_scale,carlini2017towards, dong2018boosting}. The FGSM \cite{goodfellow2014explaining} is a one-step algorithm which is proposed as a fast approximation for generating additive perturbation by moving in the direction of gradient sign of the loss function of the model. As a straightforward extension of FGSM, Kurakin et al. \cite{madry2017towards} proposed the projected gradient descent (PGD). This method iteratively applies FGSM and projects the intermediate perturbation results onto the $\epsilon-ball$ around the original image $x$. Other extenstions and variants of FGSM  have been provided such as  \cite{tramer2017ensemble, kurakin2016adversarial_ML_at_scale, madry2017towards, dong2018boosting}. In addition to FGSM and its extensions, various algorithms have been proposed to generate additive perturbations. DeepFool \cite{moosavi2016deepfool}, iteratively, computes minimal additive perturbations by taking a step towards the nearest decision boundary. Carlini \& Wagner (C\&W) \cite{carlini2017towards} generate smaller magnitude of perturbation ($L_{0}$, $L_{2}$, and $L_{\infty}$ norms) as additive perturbations using a modified loss function.  

\subsection{Unrestricted Adversarial Examples }
\label{sec:3.2}

Some of the existing adversarial attacks include ADef \cite{2018adef} which is an algorithm to construct adversarial deformation by iteratively constructing the smallest deformation to misclassify the image. Another example of such attacks is Spatially transformed attack \cite{spatially-transform2018} which solves a second-order optimization problem to find a minimal flow vector field. To deceive DNNs, each image pixel has its specific flow vector to obtain the transformation direction of the pixel. Optimization of a large number of flow vectors are inefficient in practice. SdpAdv (Spatial Distortion + Perturbation Adversary) algorithm \cite{zhao2019perturbations} applies affine transformations to craft spatial deformations (six affine matrix parameters need to be optimized) and then adds perturbations to the deforming image to generate the final UAE.
  
Previous norm-constrained AEs and UAEs only consider adding unnoticeable changes into images.
In the following, another type of UAEs is examined. These UAEs do not necessarily look exactly the same as the original images, but are still legitimate images to human eyes and can deceive the model. Such UAEs are classified as their true category using human judgment but result in misclassification in the victim model.

Brown et al. \cite{brown2017adversarial} add a small patch to original images and Eykholt et al. \cite{eykholt2018robust} print stickers on the physical stop sign in the desired positions to create an UAE. Alcorn et al. \cite{alcorn2019strike} change the poses of objects by estimating the parameters of a 3D rendering that causes the model to misclassify. Ho et al. \cite{ho2019catastrophic} create UAEs based on real-world object manipulation implemented by camera shake and pose variation. Some of the papers try to generate UAEs by changing colors and other attributes \cite{hosseini2018semantic, bhattad2019unrestricted,poursaeed2019fine}.
 
Fawzi et al. \cite{2015manitest} find the shortest path on the geometric transformation manifold, using the fast march method, to deceive the model. They examined rigid transformations (rotation, translation, and scale) and used them as a benchmark for comparing the independence of different models. Moosavi et al. \cite{moosavi2018geometric} presented the manifold algorithm, which selected the smallest transformation that deceives the model using gradient descent on the geometric transformation manifold. Then, they showed that adversarial training with this algorithm improves model robustness. Both methods \cite{2015manitest,moosavi2018geometric} have no restrictions on transformations and apply rotation, scaling, and translation to the image to the extent that the model is deceived.
Engstrom et al. \cite{engstrom2017rotation} show that natural transformations such as rotation and translation alone can be used to deceive models. They construct an UAE by solving a first-order optimization and grid search on latent space of rotation and translation parameters (using three parameters) and show that grid search is sufficient to deceive models. This method is restricted to limited number of transformations (rotation and translation) that prevents this attack from modeling real-world deformations. Proposed method exploits a similar approach but to introduce more realistic and less limited deformations to the images, it produces homographic transformations using three trainable parameters that impose restrictions on the amount of rotation, scale, translationm shear and projective warps applied to the original image.

Some papers create UAEs from scratch using a generative model. Song et al. \cite{2018_nips_unristricted_constructing} and Poursaeed et al. \cite{poursaeed2019fine} search in latent space of an Auxiliary Classifier Generative Adversarial Network (AC-GAN) and a disentangled latent representations of a style GAN, respectively, to manipulate various aspects of the image (such as color, rotation, etc.) to find the appropriate scratch to wrongly classified by model. They obtained the success rate of their attack using Amazon Turk to show that the UAEs that deceive the model are similar to the original images in human judgment.

Proposed attack does not calculate derivatives of the loss function with respect to all image pixels we are trying to optimize, rather, It calculate derivatives of the loss function with respect to just three parameters to produce a deformed UAE.

\section{Proposed Method}
\label{sec:Proposed Method}
The proposed UAE is a black-box attack which performs on arbitrary models that take an RGB or B\&W image as input. The attack relies on geometric transformations to deform the input image of a model and result in misclassification, while keeping the images recognizable to human eye. The proposed attack also aims to generate UAEs by using a minimal number of trainable parameters. In order to achieve this incentive, the proposed attack uses three trainable parameters to determine a homographic transform that when applied to input image can cause a deformation that deceives the victim model. To keep the natural appearance of images, a discriminator network is trained and used during the attack process to distinguish between original and adversarial images.
 
\subsection{Three Deceiving Parameters}

Homographic transformation is a linear transformation that relates the transformation between two planes, making it possible to retrieve the corresponding camera displacement that allows to go from the first view plane to the second one. A homographic transformation is uniquely defined by the homography matrix (H).  To determine a homography matrix that can map an image plane to another, at least four pairs of $(p,q)$ coordinates are needed; each pair containing a point coordinate in the source plane and another in the destination plane. No three points in the source points or the destination points should be in-line. This reduces the problem of finding the matrix H to finding four or more pairs of coordinates between original image and desired UAE.

To find four pairs of coordinates, one must determine eight variables $(p_1, q_1)$ $, .. ,$ $(p_4, q_4)$. However, defining limitations on these eight variables to make the final product be acceptable to human eye is a difficult task. To address this difficulty, a specific structure for mapping the source image plane to the destination plane is introduce, that both restricts the resulting transformation keeping the resulting images within wanted criteria, and reduces the trainable parameters from eight to three, to both limit some of transformations (such as rotation and translation) and improve attack generation speed.
\begin{figure}
\centering
\includegraphics[width=0.5\linewidth]{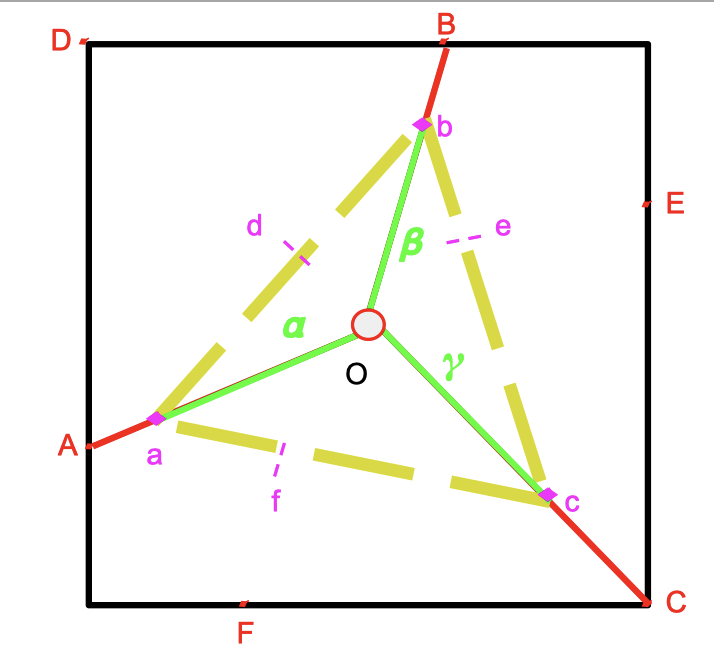}

\caption{An overview of proposed attack. Original image (black lines) is divided into three non-overlapping parts with three red borderlines based on source points $A$, $B$, and $C$, whose total area is the same. Adversarial image (yellow dashed lines) is shown with three destination points ($a$, $b$, and $c$). Using $\alpha$, $\beta$, and $\gamma$  the coordinates for three destination points ($a$, $b$, and $c$) are computed, each point being on one of the determined borderlines and each point’s distance to the center of the image being scaled by one of the scale factors  $\alpha$, $\beta$, and $\gamma$. There are three auxiliary sources ($D$, $E$, and $F$) and destination ($d$, $e$, and $f$) points to avoid having any three points in-line in the set of six destination points.}
\label{fig:2}       
\end{figure}
%
\begin{figure}
\includegraphics[width=1\linewidth]{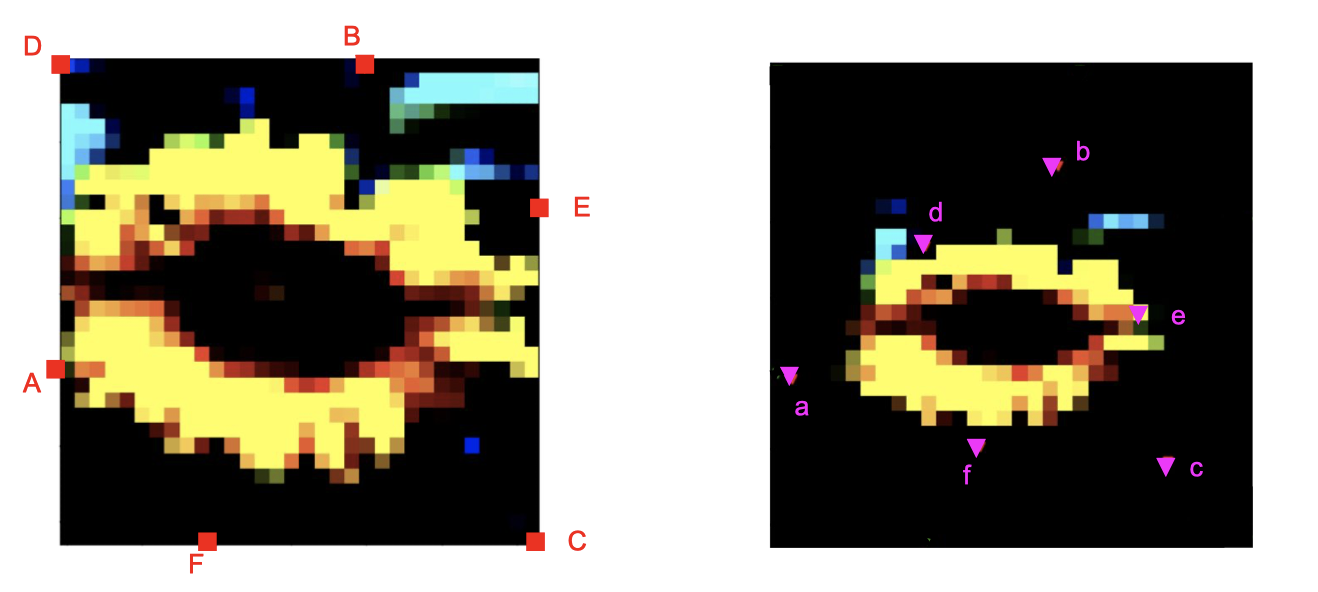}
\caption{Illustration of source and destination points on original and adversarial image of CIFAR database. (a) Original image with six source points (main and auxiliary),  shown with a small square. (b) Adversarial image with six source points (main and auxiliary), shown with a small triangle.}
\label{fig:3}       
\end{figure}
%

The proposed structure relies mainly on taking advantage of the vulnerability of CNNs to scale transformation. As shown in Figure \ref{fig:2}, in the proposed structure, a given image is divided into three non-overlapping parts that all have the same total area. After this step, the three resulting borderlines can be shown with the following Equations in the cartesian system of coordinates, where the left upper corner of the image has coordinates $(0,0)$\\
\begin{align}
&    OA: y = \frac{-h}{3w}x + \frac{2}{3}h &;& 0 \leq x \leq \frac{w}{2}
\label{eq:1}\\
&    OB: y = \frac{-3h}{w}x + 2h &;& \frac{w}{2} \leq x \leq \frac{2w}{3}
\label{eq:2}\\
&    OC: y = x &;& \frac{w}{2} \leq x \leq w
\label{eq:3}
\end{align}

where the width and height of the input image are denoted by $w$ and $h$, respectively, and the coordinates are represented by the letters $x$ and $y$. The intersection of these borderlines with the original image borders specifies three points $A$, $B$ and $C$ which are used as three source points. The coordinates for destination points which are represented by $a$, $b$ and $c$ are calculated using the above line Equations and by scaling the distance to the center of the image by the scale factors $\alpha$, $\beta$, and $\gamma$. The mapping of these three pair of points is defined by 

\begin{equation}
A \rightarrow a:(0, \frac{2h}{3}) \rightarrow (\frac{(1 - \alpha)w}{2}, \frac{(\alpha + 3)h}{6})   
\label{eq:4}
\end{equation}

\begin{equation}
B \rightarrow b:(\frac{2w}{3}, 0) \rightarrow (\frac{(3 + \beta)w}{6}, \frac{(1 - \beta)h}{2}) 
\label{eq:5}
\end{equation}

\begin{equation}
C \rightarrow c:(w, h) \rightarrow (\frac{(1 + \gamma)w}{2}, \frac{(1 + \gamma)h}{2}) 
\label{eq:6}
\end{equation}

when in Equation (\ref{eq:4}), point $A$ in the coordinates $(0, \frac{2h}{3})$ is mapped to point $a$ in the coordinates $(\frac{(1 -\alpha)w}{2}, \frac{(\alpha + 3)h}{6})$ and so are the other points in Equations (\ref{eq:5}) and (\ref{eq:6}).  Now, three variables $\alpha$, $\beta$ and $\gamma$ are determined each corresponding to a scale factor in the range of (0.2, 1). Using $\alpha$, $\beta$ and $\gamma$  then coordinates for three destination points are found, each point being on one of the determined borderlines and each point’s distance to the center of the image being scaled by one of the scale factors  $\alpha$, $\beta$ and $\gamma$. The coordinates for these points defined using $\alpha$, $\beta$, $\gamma$.\\ With having specified three source and destination points, at least one more pair of coordinates is needed to uniquely determine a homography matrix. To address this problem and also to maintain the symmetry in the structure, three more pairs of points are determined between the source and destination planes. The source points ($D$, $E$ and $F$) are chosen on original image borders and the auxiliary destination points ($d$, $e$ and $f$) are chosen on connecting lines of the three previously determined destination points, using similarity and intercept theorem. To avoid having any three points in-line in the set of destination points, a small bias (only one or two pixels shift) is added to either $p$ or $q$ in coordinates of the three added auxiliary points. These auxiliary pairs of points are shown in Figure \ref{fig:2} with a small triangle and the coordinates for them is formalized below \\
\begin{equation}
D \rightarrow d: (0,0) \rightarrow (\frac{(\beta - 3\alpha + 6)w}{12} - b, \frac{(\alpha - 3 \beta + 6)h}{12})
\end{equation}
\begin{equation}
F \rightarrow f: (\frac{w}{3}, h)\rightarrow (\frac{(\gamma - \alpha + 2)w}{4}+b, \frac{(3\gamma + 3 \alpha + 6)h}{12})    
\end{equation}
\begin{equation}
    E \rightarrow e:(w, \frac{h}{3}) \rightarrow (\frac{(3\gamma + \beta + 6)w}{12}+b, \frac{(\gamma - \beta + 2)h}{4}).
\end{equation}
Finally, by using the six pairs of source and destination points, we exploit the homographic transform presented in \cite{riba2019kornia} module that optimizes the parameters of the homography matrix H to minimize the photometric error between the original image and the transformed image using gradient descent.
\subsection{Discriminator Network}
To ensure that the resulting images look as natural as possible and remain legitimate samples for human eye, a simple discriminator network is used with three layers of convolution. This network takes as input a shuffled set of UAEs and their corresponding original images and outputs a binary indicator showing if each image is of type clean or adversary. A simple Binary Cross Entropy (BCE) loss is used to train this discriminator network in parallel to our attack scheme. This BCE loss is also utilized in the overall loss used to find the three main attack parameters $\alpha$, $\beta$ and $\gamma$.
\subsection{Training Scheme} 
The goal is to find the parameters $\alpha$, $\beta$, and $\gamma$ for each UAE so that the resulting image can fool the victim model while remaining as realistic as possible. To achieve this goal the following loss is minimized 

\begin{equation}
    \argminA_{\alpha,\beta,\gamma}{\frac{-1}{N}\sum_{o\in x_{adv}}(\sum_{c=1}^{M}y_{o.c}log(p_{o.c}) )+ \frac{-W}{2N}\sum_{o\in {x_{adv} \cup {x}}}(\sum_{a=0}^1y_{o.a}log(q_{o.a}))}
\end{equation}
where $M$ is the number of classes in the dataset, N is the number of images, $y_{o.c}$ is one if adversarial image $o$ belongs to class $c$ and zero otherwise. $p_{o.c}$ is the output probability of the victim model for a given adversarial image $o$ and class $c$. The first summation in the loss refers to a cross entropy loss used for classification results of the victim network and the second summation is the BCE loss for discriminator network. The aim is to maximize this cross entropy so that the model is deceived to misclassify the UAEs. The second part of the loss refers to the BCE loss of the discriminator. $y_{o.a}$ is one if the input image is an adversarial image and zero otherwise, $q_{o.a}$ is the probability output of model for the input being adversarial. The aim is to fool the discriminator to believe the UAEs look as natural as the original images by maximizing the BCE loss of the discriminator. The weight W is a hyper-parameter determined through validation and set to 100 for best results.  \\

\section{Experimental Results}
\label{sec:4}

The proposed method is evaluated on the six main datasets of MNIST, Fashion-MNIST, SVHN, CIFAR-10, CIFAR-100, and ImageNet which are commonly used in the image classification task. Each dataset is applied to its state-of-the-art models. The effectiveness of the attack on various architectures and datasets is measured.
\subsection{Datasets}
\label{sec:4.1}

\paragraph{MNIST}\cite{deng2012mnist}: This database of handwritten digits contains 60,000 and 10,000 images as the training and test sets, respectively. The images have been size-normalized and centered in 28x28 grayscale images.
\paragraph{Fashion-MNIST (FMNIST)} \cite{xiao2017fashion}: It consists of 60,000 training images and 10,000 testing images. Each image is a 28x28 grayscale image and corresponded to one of 10 fashion product related classes. 
\paragraph{Street View House Numbers (SVHN)} \cite{netzer2011reading}: It contains 73,257 training and 26,032  testing images classified into 1000 classes.  Each image is a 32x32 color cropped image which is obtained from house numbers in Google Street View images. The process of this digit images compared to the MNIST digit images are much more difficult because they are taken from natural scenes in the real world. 
\paragraph{CIFAR}: CIFAR-10 and CIFAR-100 datasets\cite{krizhevsky2009learning} consist of a training set of 50,000 images and a test set of 10,000 images.  Each image is a 32x32 low resolution color image and is one of 10 or 100 classes, accordingly.
\paragraph{ImageNet}\cite{krizhevsky2012imagenet}: This is one of the largest dataset for image classification. Subsets of ImageNet used in the ILSVRC2012 competitions. It contains 1.2 million training and 50,000 validation images classified into 1000 classes. 
\subsection{Settings}
\label{sec:4.2}

\paragraph{Model Architecture}: For MNIST, LENET model with two convolutional layers followed by average pooling, then two fully-connected layers and finally a softmax classifier is considered. For FMNIST, SVHN, CIFAR-10, and CIFAR-100 datasets the ResNet-18 \cite{he2016deep} and VGG-19\cite{simonyan2014very} models are employed. The ResNet-18 model was composed by five convolutional layers, one average pooling, and a fully-connected layer with a softmax at the end. VGG-19 consisted 16 convolution layers, 3 Fully connected layer, 5 MaxPool layers, and 1 softmax layer. For ImageNet,  three standard state-of-the-art models VGG-16\cite{simonyan2014very}, Inception-v3\cite{szegedy2016rethinking}, and ResNet-101\cite{he2016deep} were considered. For all these networks, the available pretrained model in Torchvision 0.8.1 was used.
As a defense models, the standard $L_{\infty}$ adversarial trained models were used according to Madry et al. \cite{madry2017towards}, on MNIST and CIFAR-10 datasets.  
All of the above architectures used in the experiments were used without any modifications.

\paragraph{Attack settings}: The proposed attack is tested on the models with the 10000 images of the MNIST, SVHN, CIFAR-10, and CIFAR-100 test sets, 26,032 images of the FMNIST, and 1000 randomly selected images from the ILSVRC2012 validation set. Also adam optimizer was used with a learning rate of 0.001 which was decayed every 20 steps by a factor of 0.1. The geometric transformations were applied using Kornia library \cite{riba2019kornia} 0.4.1 and all experiments were implemented using PyTorch 1.7.0.
 \subsection{Comparison of classification accuracy}
\label{sec:4.3}

\begin{table}
\caption{Comparison of proposed attack with other attack strategies. A lower accuracy corresponds to a stronger attack. Proposed/Mean shows the accuracy of the transformed test data created using proposed attack and the mean size of the scale factor for the three parameters $\alpha$, $\beta$, $\gamma$. Baseline column shows the accuracy of the transformed test data created where the scale factor value for the three parameters is equal to the mean obtained in the proposed method. Random column shows the accuracy of the transformed test data created when the scale factor for the three parameters is randomly selected. The remaining pixels are filled with zero-padding}
\label{tab:1}       
\begin{tabular}{c|c|cccccccccc}
\hline\noalign{\smallskip}
&  &\multicolumn{4}{c}{Attacks}   \\ 
Dataset & Model & None(\%) & Proposed(\%)/Mean & Baseline(\%) & Random(\%) \\

\hline\noalign{\smallskip} 
\multirow{1}{*}{MNIST}&LeNet\cite{lecun1998gradient} &99.06& \textbf{7.49}/0.60 & 14.63 & 36.11 \\
\hline\noalign{\smallskip}

\multirow{2}{*}{FMNIST}&VGG-19\cite{simonyan2014very} &94.4& \textbf{11.50}/0.58  &49.59 & 44.53 \\& ResNet-18 \cite{he2016deep}&92.39& 20.40/0.61  & 45.91 & \textbf{16.09}  \\
\hline\noalign{\smallskip}

\multirow{2}{*}{SVHN}&VGG-19\cite{simonyan2014very} & 94.15 & \textbf{9.19}/0.61  & 23.17 &24.12 \\& ResNet-18 \cite{he2016deep}&92.24 & \textbf{8.64}/0.58  & 25.49 & 35.62 \\
\hline\noalign{\smallskip}

\multirow{2}{*}{CIFAR10}&VGG-19\cite{simonyan2014very} & 90.28 &\textbf{15.02}/0.59  & 32.97& 26.62 \\& ResNet-18 \cite{he2016deep}&91.10 & \textbf{8.49}/0.63 & 13.72 & 24.35  \\
\hline\noalign{\smallskip}

\multirow{2}{*}{CIFAR100}&VGG-19\cite{simonyan2014very} & 70.87 &\textbf{2.12}/0.60   & 8.46 & 6.66 \\& ResNet-18 \cite{he2016deep}&72.17 & \textbf{1.63}/0.58   & 16.90 &7.65 \\
\hline\noalign{\smallskip}

\multirow{3}{*}{ImageNet}&VGG-16\cite{simonyan2014very} &74.69 &\textbf{23.98}/0.79 & 56.71 &26.87 \\ & Inception-v3\cite{szegedy2016rethinking}&71.17 &\textbf{15.75}/0.77& 46.83 & 23.22\\ & ResNet-101\cite{he2016deep}&75.40 &\textbf{20.42}/0.66& 57.10& 31.42 \\

\noalign{\smallskip}\hline
\end{tabular}
\end{table}
Table ~\ref{tab:1} demonstrates the classification accuracy of the proposed attack and other attack strategies on various datasets and models. The classification accuracy of attacks are shown as a percentage of the correctly classified test image.

The goal of the proposed method is to find the $\alpha$, $\beta$, and $\gamma$ parameters for each UAE, so that the resulting image can deceive the victim model with minimal deformation while remaining as realistic as possible. Therefore, after constructing an UAE for the test image, the mean of $\alpha$, $\beta$, and $\gamma$ values are calculated separately, and reported them as Mean in Table \ref{tab:1}.  For example, in the ImageNet dataset, the mean value are $\alpha$ = 0.93, $\beta$ = 0.79, and $\gamma$ =0.66. Therefore mean, of ($\alpha$, $\beta$, $\gamma$) is 0.79. Naturally, mean indicates the deformation rate of the adversarial image. The higher the mean, the lower the required deformation rate for the attack. 

Given that the model accuracy is affected when different scale factors are applied to the input image, second experiment is designed called baseline. In baseline, $\alpha$ = $\beta$ = $\gamma$ = $mean$ is set and the accuracy is reported in the baseline column of Table \ref{tab:1}. Comparison of baseline with the proposed attack shows that the proposed attack with effective selection of $\alpha$, $\beta$, and $\gamma$ can reduce the accuracy by about 21\% compared to the baseline.

The effect of different scale factors on the input image has also been examined. These results can be seen in Appendix A (Table \ref{tab:5}. Nine different scale factors in the range of 0.2 to 1 are applied to the input image, provided that in each scale factor change, three variables $\alpha$, $\beta$ and $\gamma$ receive the same amount of scale change. 
To make the image size equal to the size of the input image, the empty space is filled with zero-padding (black pixels). The accuracy of the models on these settings is given in in Appendix B (Table \ref{tab:6}). 
 
\begin{figure*}
    \centering
\captionsetup[subfigure]{labelformat=empty}
\captionsetup[subfloat]{position=bottom}
\captionsetup[subfigure]{font=scriptsize,labelfont=small}

    \caption*{\ \ \ \ \  Original \ \ \ \ \ \      Adversary  \ \ \ \ \ \ \ \ \ \    Original  \ \ \ \ \ \     Adversary  \ \ \ \ \ \ \ \ \ \     Original    \ \ \ \ \ \     Adversary  \ \ \  \ }

    \subfloat[5]{{\includegraphics[width=1.6cm]{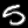} }}
    \subfloat[9]{{\includegraphics[width=1.6cm]{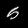} }}
     \qquad
    \subfloat[7]{{\includegraphics[width=1.6cm]{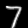} }}
    \subfloat[3]{{\includegraphics[width=1.6cm]{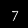} }}
     \qquad
     \subfloat[9]{{\includegraphics[width=1.6cm]{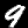} }}
     \subfloat[3]{{\includegraphics[width=1.6cm]{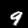} }}
     \qquad
     
    \subfloat[Shirt]{{\includegraphics[width=1.6cm]{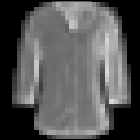} }}
    \subfloat[Bag]{{\includegraphics[width=1.6cm]{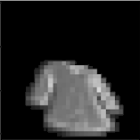} }}
     \qquad
    \subfloat[Trouser]{{\includegraphics[width=1.6cm]{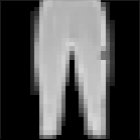} }}
    \subfloat[Sandals]{{\includegraphics[width=1.6cm]{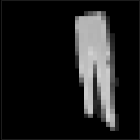} }}
     \qquad
     \subfloat[Bag]{{\includegraphics[width=1.6cm]{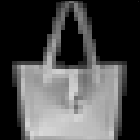} }}
     \subfloat[Sandals]{{\includegraphics[width=1.6cm]{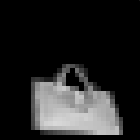} }}
     \qquad
     
      \subfloat[3]{{\includegraphics[width=1.6cm]{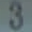} }}
    \subfloat[9]{{\includegraphics[width=1.6cm]{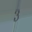} }}
     \qquad
    \subfloat[5]{{\includegraphics[width=1.6cm]{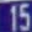} }}
    \subfloat[3]{{\includegraphics[width=1.6cm]{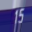} }}
     \qquad
     \subfloat[5]{{\includegraphics[width=1.6cm]{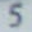} }}
     \subfloat[7]{{\includegraphics[width=1.6cm]{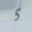} }}
     \qquad
     
    \subfloat[Horse]{{\includegraphics[width=1.6cm]{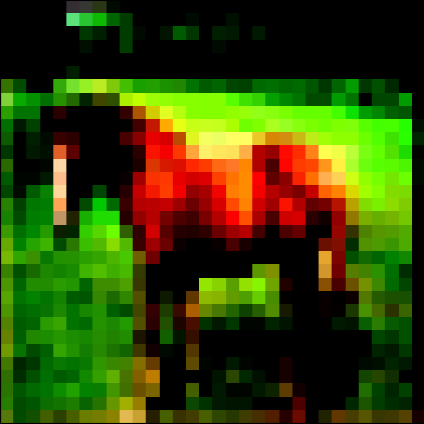} }}
    \subfloat[Deer]{{\includegraphics[width=1.6cm]{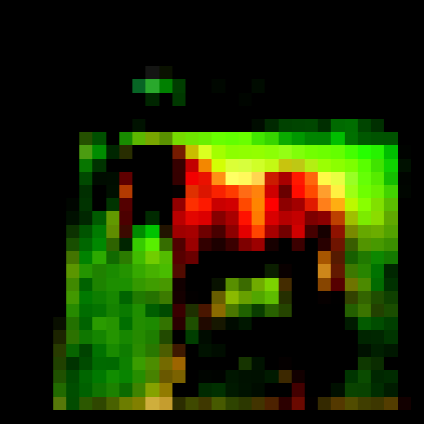} }}
     \qquad
    \subfloat[Dog]{{\includegraphics[width=1.6cm]{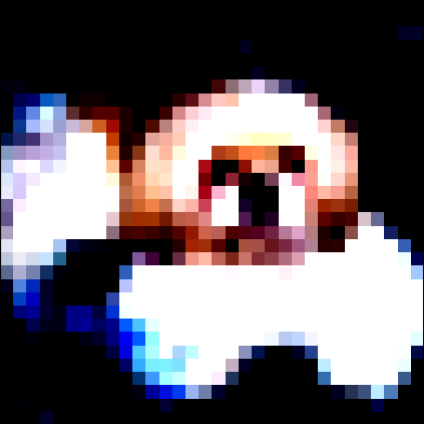} }}
    \subfloat[Cat]{{\includegraphics[width=1.6cm]{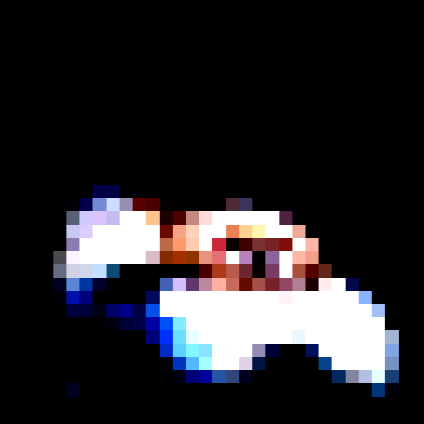} }}
     \qquad
     \subfloat[Truck]{{\includegraphics[width=1.6cm]{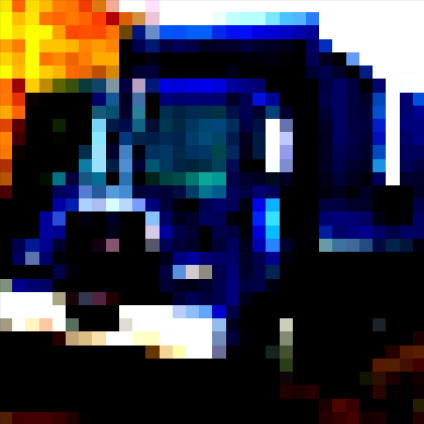} }}
     \subfloat[Automobile]{{\includegraphics[width=1.6cm]{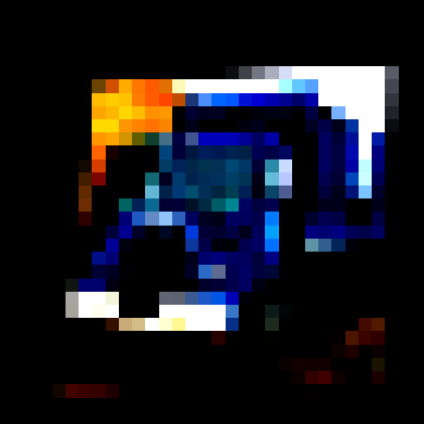} }}
     \qquad
     
      \subfloat[Iguana]{{\includegraphics[width=1.6cm]{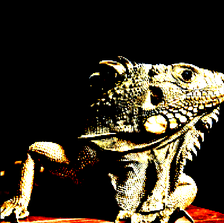} }}
    \subfloat[Weevil]{{\includegraphics[width=1.6cm]{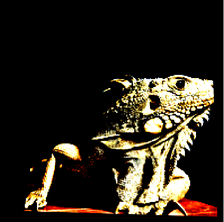} }}
     \qquad
    \subfloat[White Shark]{{\includegraphics[width=1.6cm]{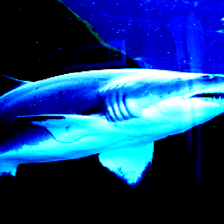} }}
    \subfloat[Gray Wolf]{{\includegraphics[width=1.6cm]{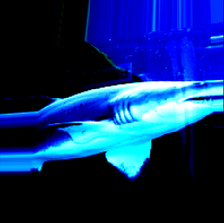} }}
     \qquad
     \subfloat[American \ \ Eagle]{{\includegraphics[width=1.6cm]{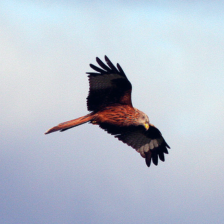} }}
     \subfloat[Red-breasted \ \ merganser]{{\includegraphics[width=1.6cm]{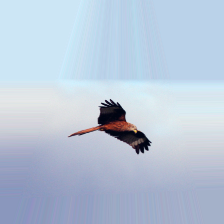} }}
     \qquad
     
    \subfloat[Tibetan \ \ \ Terrier]{{\includegraphics[width=1.6cm]{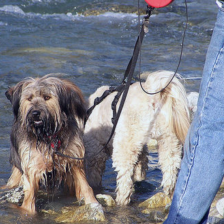} }}
    \subfloat[Briard]{{\includegraphics[width=1.6cm]{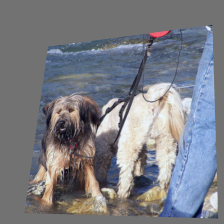} }}
     \qquad
    \subfloat[Eft]{{\includegraphics[width=1.6cm]{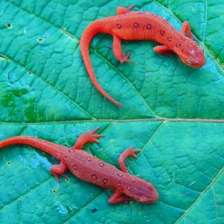} }}
    \subfloat[Starfish]{{\includegraphics[width=1.6cm]{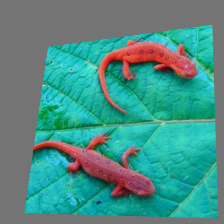} }}
     \qquad
     \subfloat[Dingo]{{\includegraphics[width=1.6cm]{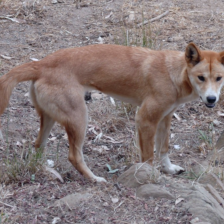} }}
     \subfloat[Saluki]{{\includegraphics[width=1.6cm]{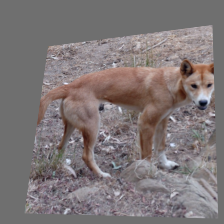} }}
     \qquad

 \caption{Random examples of successful urestricted adversarial examples created with proposed attack for MNIST, FMNIST, SVHN, CIFAR, and ImageNet datasets, respectively. [Odd columns: Original images with labels. Even columns: Adversarial examples with estimated labels. Remaining pixels in rows 3 and 6 are filled with border-extrapolation and filled with zero-padding in other columns.]}
  \label{fig:4}       
     \end{figure*}

To show the efficiency of the proposed method, the parameters $\alpha$, $\beta$, and $\gamma$ are found randomly and the model accuracy is obtained on the resulting image. It is observed that the random mode has an accuracy less than the baseline mode and the accuracy of the proposed method is in average 13\% lower than this mode, which demonstrates that the proposed method can effectively increase the vulnerability of models to scale deformation.
To have an image with the same size as the input image, the empty space is filled with both zero-padding (black pixels) and border-extrapolation (border pixels are extrapolated). The accuracy of the models, listed in Table \ref{tab:1}, is related to zero-padding. Accuracy of the models related to border-extrapolation is reported in Table \ref{tab:6} in Appendix B.
Examples of successful UAEs created with the proposed attack are visualized in Figure \ref{fig:4}.


Given that the proposed attack is an UAE and uses only three parameters to deform the image, AdvRotTran implemented in \cite{2019exploring} is the most related attack to this work among the many related attack methods discussed in Section ~\ref{sec:Related Work}. Because AdvRotTran attack, as well as proposed attack, is an UAE that makes noticeable changes with considering spatial deformation in the UAEs. It also uses three parameters to create deformation (including rotation and translation only) in the image. Translations and rotations are restricted to 10\% of the image pixels and 30 degree, respectively. AdvRotTran uses a non-differentiable method to select UAEs from space of allowed deformations, while the proposed attack is a differentiable method, which is simpler and more flexible to train. In the last column, Table \ref{tab:2} shows the accuracy of AdvRotTran with only three parameters. The proposed method is more deformation than AdvRotTran. In addition to rotation and translation, it can cover scale, shear and projective warps. It also reduces accuracy by an average of about 13\%.


\begin{table}
\centering
\caption{Comparison of proposed attack with AdvRotTran\cite{2019exploring} attack. A lower accuracy corresponds to a stronger attack. AdvRotTran column: accuracy of  transformed dataset created using three parameters used in AdvRotTran. }
\label{tab:2}       
\begin{tabular}{c|c|cccccccccc}
\hline\noalign{\smallskip}
&  &\multicolumn{3}{c}{Attacks}   \\ 
Dataset & Model & None(\%) & Proposed(\%) & AdvRotTran(\%)\cite{2019exploring}\\

\hline\noalign{\smallskip} 
\multirow{1}{*}{MNIST}&LeNet\cite{lecun1998gradient} &99.06& \textbf{7.49} & 26.02\\
\hline\noalign{\smallskip} 

\multirow{1}{*}{FMNIST}& ResNet-18 \cite{he2016deep}&92.39& \textbf{20.40}  & 46.73 \\
\hline\noalign{\smallskip}

\multirow{1}{*}{CIFAR10}& ResNet-18 \cite{he2016deep}&91.10 & 8.49 & \textbf{2.80} \\
\noalign{\smallskip}\hline
\end{tabular}
\end{table}
\subsection{Human Evaluation}
\label{sec:4.4}
In contrast to adding imperceptible changes into images that makes the adversarial example to look exactly the same as the original images, UAEs with perceptible changes must be verified by human evaluation. To show proposed UAEs are classified as their true category using human judgement, and are still legitimate images to the human eyes, 100 images from MNIST and SVHN datasets (10 per class) are randomly selected to generate their UAEs pair. Each UAE is assigned to 11 volunteers, and the majority of votes are considered as the label. The volunteer must choose label of the image between all classes (from 0 to 9) and X, whith X meaning that the image does not look close to any of the 10 classes. In Table \ref{tab:3}, it can be seen that 98\% and 89\% of majority votes match the correct classes in MNIST and SVHN datasets, respectively. Success rate of each class can also be seen in Table \ref{tab:3}. Only these datasets are used,  because they are easier and less ambiguous for human judgment, compared to the more complex labels of some other datasets, such as FMNIST\cite{xiao2017fashion}, CIFAR\cite{krizhevsky2009learning}, and ImageNet\cite{krizhevsky2012imagenet}. 


\begin{figure}
\includegraphics[width=0.5\linewidth]{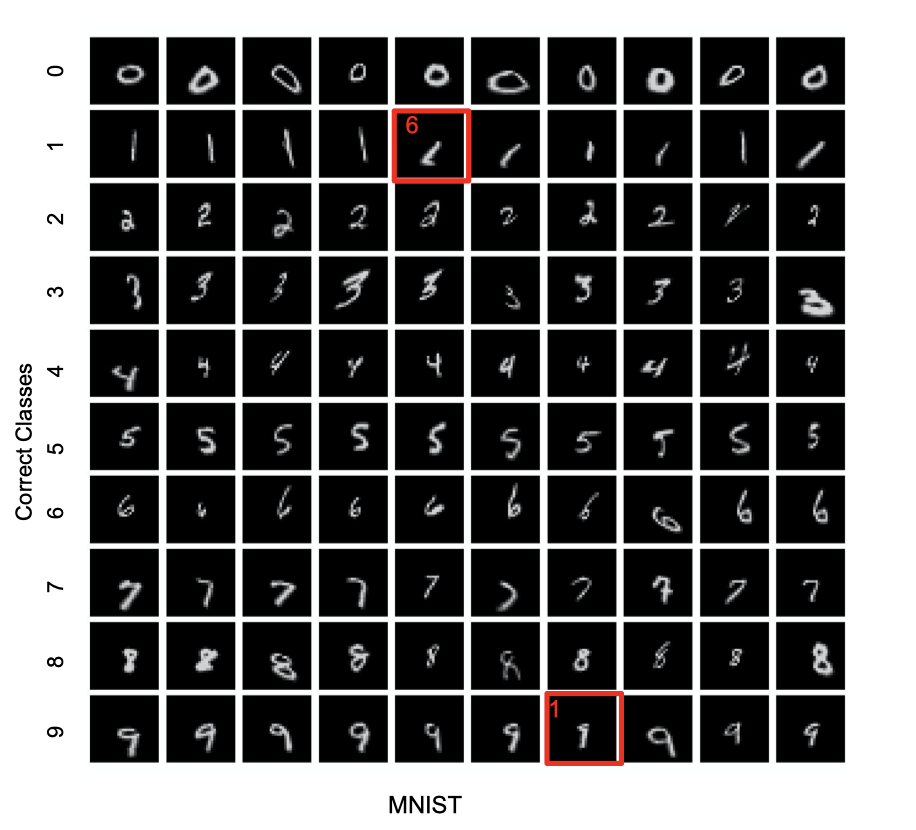}
\includegraphics[width=0.5\linewidth]{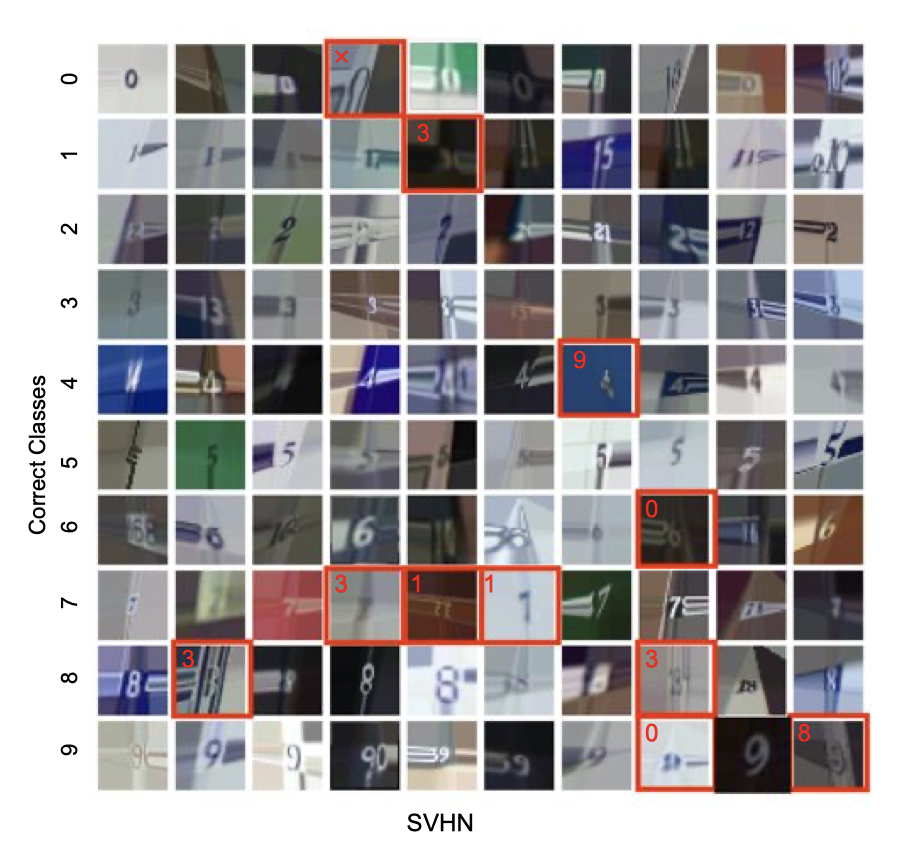}

\caption{Unrestricted adversarial examples from MNIST and SVHN datasets. Each row corresponds to adversarial examples generated by a class (from 0 to 9). These images were randomly given to 11 volunteers. Red boxes and red text inside them show the label obtained by a majority vote that failed to assign the correct class. As can be seen 98\% and 89\% of majority votes match to correct classes in MNIST and SVHN datasets, respectively.}

\label{fig:5}       
\end{figure}
%

\begin{table}
\caption{ Success rates (\%) in term of Human Evaluation. First row: class label of classes. Second and third rows: success rates (\%) of proposed attack for each class and total of them, based on human evaluation.}
\label{tab:3}       
\begin{tabular}{c|c|c|c|c|c|c|c|c|c|c|c}
\hline\noalign{\smallskip}

Dataset/Success rate(\%) & 0 & 1 & 2 & 3 & 4 & 5 & 6 & 7 & 8 & 9 & total \\

\noalign{\smallskip}\hline\noalign{\smallskip}
 MNIST & 100 & 90 & 100 & 100 & 100 & 100 & 100 & 100 &  100 & 90 & 98\\
SVHN & 90 & 90 & 100 & 100 & 90 & 100 & 90 & 70 &  80 & 80 & 89\\
\noalign{\smallskip}\hline
\end{tabular}
\end{table}


\subsection{Adversarial Training}
\label{sec:4.5}

It has been shown that adversarial training increases the robustness of the model by injecting AEs into the training data. Therefore, LeNet model are trained on a mixture of original images and proposed UAEs on MNIST dataset. Specifically, for each iteration, first, a batch of original images are fed to the model, then the corresponding batch of proposed UAEs is generated and fed alternatively. In Figure \ref{fig:3}, the evaluation results of adversarial training are compared with original training approach. Where original training means the model is trained only on original images. Two approaches (original and adversarial training) are tested on images that have been randomly transformed to demonstrate the power of using the proposed UAEs on model robustness.

The numbers on the horizontal axis indicate the range in which $\alpha$, $\beta$ ,and $\gamma$ have been selected and applied to the randomly selected images. It is observed that the longer the transform interval of the test image, the greater the distance between the two approaches. For instance, when the transformation limit is between 0.8-1, the accuracy of the adversarial training approach is 94.42\% and the accuracy of the original training approach is 78.62\% (the distance between the two approaches here is 15.8\%). If the limit is set to a larger range such as 0.6-1, it is observed that the accuracy of the adversarial training approach and the accuracy of the original training approach, decreases to 93.62\% and 65.49\%, respectively, (the distance between the two approaches here has reached 28.13\%, meaning that the bigger range resulted in an increase of 12.33\%). In addition, it can be seen that adversarial training has been able to achieve good accuracy against randomly transformed images even when allowing large intervals for $\alpha$, $\beta$, and $\gamma$. Experiments show the adversarial training with proposed UAEs has increased model robustness on the randomly transformed image.

\begin{figure}
\centering
\includegraphics[width=0.8\linewidth]{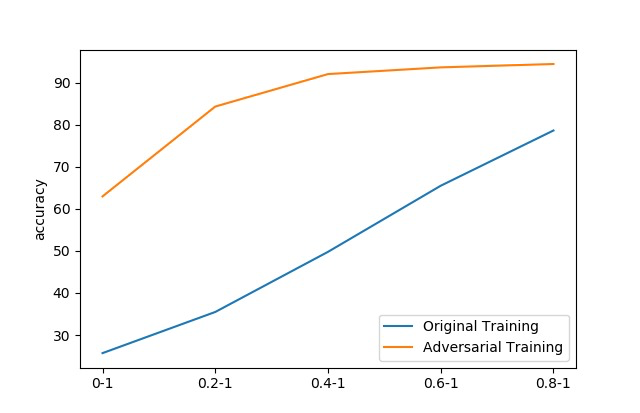}
\caption{Adversarial training and original training results of LeNet model on MNIST dataset. Two approaches (original and adversarial training) are tested on images that were randomly transformed to demonstrate the power of using the proposed unrestricted adversarial examples on model robustness. Numbers on horizontal axis indicate the range in which $\alpha$, $\beta$, and $\gamma$ have been selected to be applied to the randomly selected test images.}
\label{fig:3}       
\end{figure}
%

\subsection{Transferability}
\label{sec:4.6}

Enemy accuracy in each target model when producing UAE The UAE was fed to it by the source model.

One of the most important features of an attack is its ability to to be generalized to different models. To test how proposed attack can be transferable, three pre-trained models VGG-16\cite{simonyan2014very}, Inception-v3\cite{szegedy2016rethinking}, and ResNet-101\cite{he2016deep} are used on the same Imagenet dataset and are reported in Table \ref{tab:4} the adversary accuracy of each target model when the UAEs generated by the source model were fed to it. The results in Table \ref{tab:4} show that the proposed attack can be generalized between different models, which means the UAEs can be a threat in a black box scenario. Results for other datasets such as CIFAR10 are presented in Table \ref{tab:7} of Appendix C.

\begin{table}
\centering
\caption{Transferability of proposed unrestricted adversarial examples on ImageNet. Numbers indicate adversary accuracy of proposed attack between pairs of models for 1000 random images from ImageNet dataset. Rows and columns:  source and target models, respectively.}
\label{tab:4}       
\begin{tabular}{c|c|c|c}
\hline\noalign{\smallskip}
Model & VGG-16\cite{simonyan2014very} & Inception-v3\cite{szegedy2016rethinking} & ResNet-101\cite{he2016deep}\\
    
\hline\noalign{\smallskip} 
 VGG-16\cite{simonyan2014very} & 22.98 & 32.18 & 40.22 \\

\hline\noalign{\smallskip} 
 Inception-v3\cite{szegedy2016rethinking} & 37.02 & 15.75 & 40.66 \\
 
 \hline\noalign{\smallskip} 
 ResNet-101\cite{he2016deep} & 35.00 & 42.39 & 20.42 \\

\noalign{\smallskip}\hline
\end{tabular}
\end{table}


\section{Conclusion}
\label{sec:5}

A new UAEs was proposed to deceive models without affecting human predictions. The proposed attack generates UAEs by applying limited spatial deformation on original images, including scaling, rotation, shear, and translation, by using only three trainable parameters. By selecting three points on the original image and based on the location of these three points, a homographic transformation is built and applied to the original image to make an unrestricted adversarial example. Experimental results show that the proposed adversarial examples have an average of 99\% success rate in terms of human evaluation on the MNIST and SVHN datasets and can reduce the model accuracy by an average of 73\% in six datasets of MNIST, FMNIST, SVHN, CIFAR10, CIFAR100, and, ImageNet. Also, adversarial training using the proposed attack can improve the model robustness against a randomly transformed image.
Future work can explore using four or more points of the original image to generate more varied and less limited transformations.\\

\textbf{Acknowledgement.} The authors would like to thank 
Dr. Seyed-Mohsen Moosavi-Dezfooli for the helpful discussions.


%
%

\clearpage
\bibliographystyle{spbasic}      
\bibliography{myref}   

\clearpage
\appendix
\counterwithin{table}{section}

\label{sec:6}

\renewcommand{\theequation}{A-\arabic{equation}}    
  \setcounter{equation}{0}  
  \section*{Appendix A}  
  \setcounter{table}{0} \renewcommand{\thetable}{A.\arabic{table}}


Nine different scale factors in the range of 0.2 to 1 are applied to the input image, provided that in each scale factor change, three variables $\alpha$, $\beta$, and $\gamma$ receive the same amount of scale change. To make the image size equal to the size of the input image, the empty space is filled with zero-padding (black pixels). The accuracy of the models on these settings is given in Table \ref{tab:5}. 
As expected, the smaller the factor scale, the lower the accuracy of the models. According to the results listed in Table \ref{tab:5}, with each step of the scale factor reduction, the accuracy of the models decreases by approximately 8\%. 

\begin{landscape}
\begin{table}[hbt!]
\caption{Accuracy of models in various scale factors of input image. Constant C to three variables  $\alpha$, $\beta$, and $\gamma$ is set as $\alpha$ = $\beta$ = $\gamma$ = C. Assigned C to nine different values in the range of (0.2, 1), which are shown in Columns 3 to 9. or example Third column shows the accuracy of the models when  $\alpha$ = $\beta$ = $\gamma$ = 0.2. }
\label{tab:5}       
\begin{tabular}{c|c|cccccccccc}
\hline\noalign{\smallskip}
Dataset & Model & 0.2(\%) & 0.3(\%) & 0.4(\%) & 0.5(\%) & 0.6(\%) & 0.7(\%) & 0.8(\%) & 0.9(\%) & 1(\%)\\

\hline\noalign{\smallskip} 
\multirow{1}{*}{MNIST}&LeNet\cite{lecun1998gradient} & 14.63 & 19.51 & 22.55 & 28.21 & 37.83 & 51.72 & 66.71 & 78.90 & 86.41\\
\hline\noalign{\smallskip}

\multirow{2}{*}{FMNIST}&VGG-19\cite{simonyan2014very} &13.84 & 21.60  & 28.27 & 39.53 & 54.48 & 69.08 & 79.96 & 86.52 & 89.74\\& ResNet-18 \cite{he2016deep}& 16.09 & 22.06 & 30.91 &44.52 &60.39 &76.35 & 85.35 & 89.71 & 91.70 \\
\hline\noalign{\smallskip}

\multirow{2}{*}{SVHN}&VGG-19\cite{simonyan2014very} & 130.9  & 16.97 & 19.97 & 29.55 & 39.43 & 51.87 & 64.80 & 79.32 & 87.74 \\& ResNet-18 \cite{he2016deep}& 14.58 & 13.75 & 18.78 & 34.15 & 46.99 & 58.44 & 69.81 & 82.17 & 89.97   \\
\hline\noalign{\smallskip}

\multirow{2}{*}{CIFAR10}&VGG-19\cite{simonyan2014very} & 12.35 & 14.72 & 16.28 & 21.08 & 29.00 & 38.64 & 50.55 & 61.15 & 72.25 \\& ResNet-18 \cite{he2016deep}& 12.30  & 14.16 & 18.27 & 23.60 & 31.08 & 40.34 & 52.64 & 64.50 & 74.10 &  \\
\hline\noalign{\smallskip}

\multirow{2}{*}{CIFAR100}&VGG-19\cite{simonyan2014very} & 1.81 & 2.41 & 3.17 & 4.75 & 7.94 & 12.71 & 22.77 & 35.24 & 49.92 & \\& ResNet-18 \cite{he2016deep}&  2.10 & 2.70 & 3.51 & 4.95 & 8.88 & 15.46 & 26.22 & 38.30 & 52.38 \\
\hline\noalign{\smallskip}

\multirow{2}{*}{ImageNet}&VGG-16\cite{simonyan2014very} & 1.03 & 5.14 &18.43 &29.14  & 37.94 & 46.07 & 51.86 & 57.74 & 62.71\\ & Inceptiion-v3\cite{szegedy2016rethinking}& 0.91 & 5.21 & 13.05 & 25.21 & 35.22& 44.93 & 51.01 & 55.56 & 60.47\\ & ResNet-101\cite{he2016deep}&2.09 &10.27& 25.59 & 39.42 & 51.87 & 59.98 & 64.13 & 66.30 & 69.41 \\

\noalign{\smallskip}\hline
\end{tabular}
\end{table}
\end{landscape}
\clearpage

\renewcommand{\theequation}{A-\arabic{equation}}    
  \setcounter{equation}{0}  
  \section*{Appendix B}  
  \setcounter{table}{0} \renewcommand{\thetable}{B.\arabic{table}}


To make the image size equal to the size of the input image, the empty space is filled with border-extrapolation (border pixels are extrapolated). The classification accuracy of the state-of-the-arts models on SVHN and ImageNet datasets, listed in Table \ref{tab:6}, is related to border-extrapolation.
\begin{table}[hbt!]
\caption{Comparison of proposed attack with other attack strategies. A lower accuracy corresponds to a stronger attack. Proposed/Mean shows the accuracy of the transformed test data created using proposed attack and the mean size of the scale factor for the three parameters $\alpha$, $\beta$, $\gamma$. The baseline column shows the accuracy of the transformed test data created where the scale factor value for the three parameters is equal to the mean obtained in the proposed method. The random column shows the accuracy of the transformed test data created when the scale factor for the three parameters is randomly selected. The remaining pixels are filled with border-extrapolation}
\label{tab:6}       
\begin{tabular}{c|c|cccccccccc}
\hline\noalign{\smallskip}
&  &\multicolumn{4}{c}{Attacks}   \\ 
Dataset & Model & None(\%) & Proposed(\%)/Mean & Baseline(\%) & random(\%) \\

\hline\noalign{\smallskip} 

\multirow{2}{*}{SVHN}&VGG-19\cite{simonyan2014very} & 94.15 & \textbf{6.74}/0.60  & 34.42 &34.83 \\& ResNet-18 \cite{he2016deep}&92.24 & \textbf{6.91}/0.53  & 29.73 & 37.45 \\
\hline\noalign{\smallskip}

\multirow{3}{*}{ImageNet}&VGG-19\cite{simonyan2014very} &74.69 &\textbf{17.95}/0.61 & 33.30 & 55.37\\ & Inceptiion-v3\cite{szegedy2016rethinking}&71.17 &\textbf{15.47}/0.63 & 37.20 & 51.67 \\ & ResNet-101\cite{he2016deep}&75.40 &\textbf{23.41}/0.68 & 44.07 & 59.78 \\

\noalign{\smallskip}\hline
\end{tabular}
\end{table}

\renewcommand{\theequation}{A-\arabic{equation}}    
  \setcounter{equation}{0}  
  \section*{Appendix C}  
    \setcounter{table}{0} \renewcommand{\thetable}{C.\arabic{table}}


The results in Table \ref{tab:7} show that the proposed attack can be generalized between different state-of-the-art models on the same CIFAR10 dataset. In Table \ref{tab:7}, the adversary accuracy of each target model is seen when the UAEs generated by the source model were fed to it.
\begin{table}[hbt!]
\centering
\caption{Transferability of proposed unrestricted adversarial examples on CIFAR10 dataset. Numbers indicate the adversary accuracy of proposed attack between pairs of models for test images from CIFAR10 dataset. Rows and columns: source and target models, respectively.}
\label{tab:7}       
\begin{tabular}{c|c|c}
\hline\noalign{\smallskip}
Model & VGG-19\cite{simonyan2014very} & ResNet-18\cite{he2016deep}\\

\hline\noalign{\smallskip} 
 VGG-19\cite{simonyan2014very} & 15.02 & 21.34  \\

 \hline\noalign{\smallskip} 
 ResNet-18\cite{he2016deep} & 12.11 & 8.49  \\

\noalign{\smallskip}\hline
\end{tabular}
\end{table}

\end{document}